\documentclass{article}
\usepackage{ijcai25}

\usepackage{times}
\usepackage{soul}
\usepackage{url}
\usepackage{graphicx}

\usepackage[hidelinks]{hyperref}
\usepackage[utf8]{inputenc}
\usepackage[small]{caption}
\usepackage{amsmath}
\usepackage{booktabs}
\usepackage{algorithm}
\usepackage{algorithmic}
\usepackage[switch]{lineno}
\usepackage{hyperref}

\usepackage{graphicx}
\usepackage{tabularray}
\usepackage{tabularx}
\usepackage{multirow} 
\usepackage{pgfplots}
\usepackage{pgfkeys}
\usepackage{pgfplotstable}
\usepackage{xfp}
\usepackage{subcaption}
\usepackage{array}

\usepgfplotslibrary{statistics}

\title{The Graph's Apprentice: Teaching an LLM Low-Level Knowledge for Circuit Quality Estimation}

\author{Reza Moravej \thanks{Correspondence to \texttt{reza.moravej@huawei.com}} , Saurabh Bodhe, Zhanguang Zhang, Didier Chételat, Dimitrios Tsaras, Yingxue Zhang, Hui-Ling Zhen, Jianye Hao, Mingxuan Yuan \affiliations
Huawei Noah's Ark Lab
}

\begin{document}

\maketitle
\begin{abstract}
Logic synthesis is a crucial phase in the circuit design process, responsible for transforming hardware description language (HDL) designs into optimized netlists. However, traditional logic synthesis methods are computationally intensive, restricting their iterative use in refining chip designs. Recent advancements in large language models (LLMs), particularly those fine-tuned on programming languages, present a promising alternative. This work proposes augmenting LLMs with predictor networks trained to estimate circuit quality directly from HDL code. To enhance performance, the model is regularized using embeddings from graph neural networks (GNNs) trained on Look-Up Table (LUT) graphs, thereby incorporating lower-level circuit insights. The proposed method demonstrates superior performance compared to existing graph-based RTL-level estimation techniques on the established benchmark OpenABCD, while providing instant feedback on HDL code quality.
\end{abstract}
\section{Introduction} \label{sec:introduction}

Rapid technological advancements in computing power has taken an increasingly important role in the past decades in driving scientific research in biology , chemistry , physics and especially artificial intelligence, where it has been estimated that at least half of all performance gains in the past ten years have stemmed from hardware improvements alone \cite{dorner2021measuring,karpathy2022deep,erdil2022algorithmic}. This ever-rising demand for compute power means that efficient and effective electronic chip design has become increasingly critical.

Modern electronic chip design is a complex, multi-stage endeavor that begins with a chip architect specifying the digital circuit's functionality in a Hardware Description Language (HDL), such as Verilog \cite{thomas2008verilog} or VHDL \cite{coelho2012vhdl}. This HDL code is then subjected to a series of transformations and optimizations, ultimately yielding a physical circuit design that can be manufactured \cite{lameres2023introduction}. 
The quality of the resulting circuits are usually measured using physical characteristics only available in the later stages, such as circuit area or delay \cite{brayton2010abc}. However, the computational cost of logic synthesis makes the iterative improvement according to the resulting circuit quality metrics prohibitively expensive.  Optimizations are best made early on in this pipeline, ideally at the RTL level, so as to leave maximal flexibility in circuit design. Thus, efficient feedback methods that can estimate the quality of results of the HDL, can improve the resulting circuit and reduce the overall chip design time. 



This discrepancy has led to interest in using artificial intelligence methods in the circuit design process \cite{huang2021machine}. In this literature, machine learning models are trained to provide feedback on HDL code without running the actual logic synthesis process. This is done using supervised learning on a training set of circuits for which logic synthesis has been run and from which quality-of-result (QoR) metrics such as circuit area and delay have been computed. Although this approach seems straightforward, finding an  representation of the RTL code appropriate for machine learning models has proven a challenge. The few works that have approached this topic did so by extracting graphical information about the code and using hand-designed statistics of those graphs as features \cite{zhou2019primal,sengupta2022iccad,fang2023masterrtl}. Despite encouraging results, the performance of these methods has ultimately been limited by the relatively shallow understanding of the semantics of the code that these statistics can provide.

\begin{figure}[t]
    \centering
    \includegraphics[clip, trim={20pt, 0pt, 20pt, 0pt}, width=\linewidth]{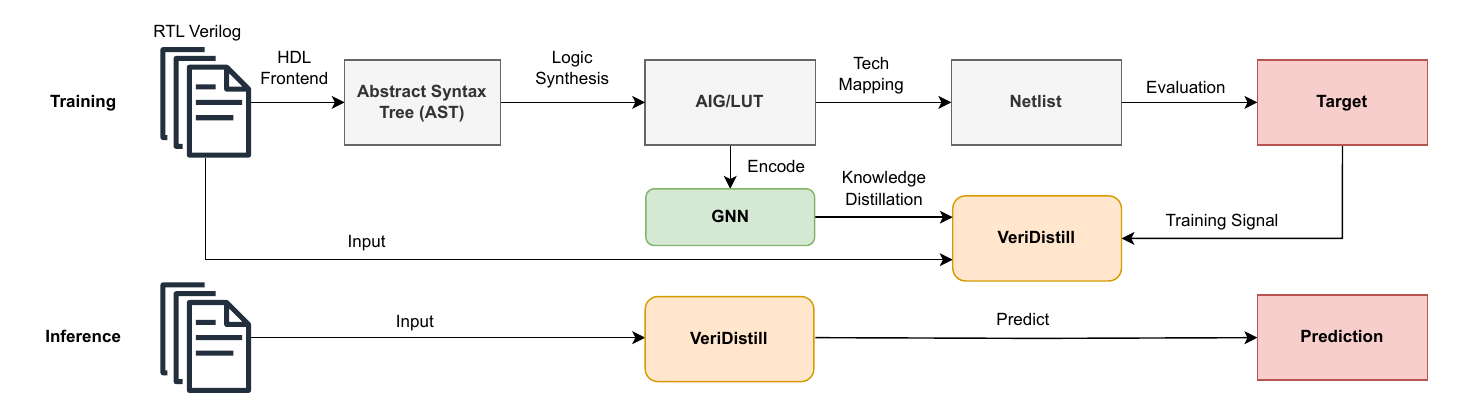}
    \caption{Overview of the training and inference pipeline. During training, LUT graphs, and the area/delay labels are used to train the model. During inference, only the source Verilog is required to generate the post-synthesis area/delay prediction.}
    \label{fig:overview}
\end{figure}
Recently, Large Language Models fine-tuned on code, such as Code-T5~\cite{wang2021codet5}, Codex~\cite{chen2021evaluating}, CodeGen~\cite{nijkamp2022codegen}, CodeLlama~\cite{roziere2023code} and DeepSeek-Coder~\cite{guo2024deepseek},  have proven to be remarkably successful on a wide range of tasks \cite{zheng2023survey}, most notably as code assistants such as Github Copilot\footnote{https://github.com/features/copilot}. This raises the question as to whether their internal representations could be used as inputs to machine learning to predict circuit quality estimates.



In this work, we propose to feed Verilog code to the state-of-the-art Large Language  Models, and train an inexpensive decoder neural network that uses the LLM's hidden states as features to predict area and delay. In addition, and critically, we regularize this decoder to encourage its embeddings to resemble those of a graph neural network model trained on Look-Up Table (LUT) graph, an intermediate representation used during the logic synthesis process. The resulting decoder is shown to strongly outperform state-of-the-art baselines in RTL-level circuit quality estimation, while keeping training and inference costs practical.

Our work makes the following main contributions:
\begin{enumerate}
\item We develop the first truly end-to-end machine learning model in the literature, named VeriDistill, which can take raw Verilog code, without any preprocessing, and produce accurate estimates of circuit area/delay metrics.
\item Moreover, we apply during training a novel knowledge distillation method which allows to transfer low-level insights about the circuit, in the form of LUT graphs, back into the machine learning predictor model.
\item We demonstrate through experiments that the combination of those two elements outperforms previous state-of-the-art baselines in a large-scale Verilog dataset and enhances the model's ability to transfer to out-of-distribution data.
\item Finally, we also demonstrate that both using LLM representations and the knowledge distillation are essential, in that removing any one of these components brings the performance back below the previous baselines.
\end{enumerate}

The remainder of this paper is structured as follows. Section \ref{sec:background} provides an overview of the relevant literature and background information. In Section \ref{sec:method}, we present a detailed description of our proposed methodology, including its key components and underlying assumptions. The efficacy of our approach is then demonstrated through a series of experiments, which are reported in Section \ref{sec:experiments}. Finally, Section \ref{sec:conclusion} summarizes our main findings, discusses their implications, and outlines potential avenues for future research.

\section{Related Work} \label{sec:background}

\subsection{Quality-of-Result Prediction from HDL Code}

Closest to ours is the work of \cite{sengupta2022iccad}. Their approach consists in computing the Abstract Syntax Tree (AST) induced by Verilog code, and extracting from this free vector- and graph-based features. They then train several machine learning models to predict from these features the total negative slack and dynamic power of the circuit. Among all the models evaluated, the XGBoost Regressor performs best and achieves 95\% R2-score. The analysis was however limited to different runs of a single circuit and it is not clear how the performance would generalize to different circuits. Since the Abstract Syntax Tree is essentially the raw Verilog code with extra syntactic information, which can be obtained at little cost at inference time by a grammar parser, we include it (along with variants) as baselines in our experimental section.

Further related is the work of \cite{fang2023masterrtl} and \cite{fang2024transferable}. They propose to process Verilog code into a new representation called Simple Operator Graph (SOG), and test several machine learning models (Transformers, Random Forests, Graph Neural Networks and XGBoost regressors) to predict path delay, module-level power and combinatorial area. Although achieving promising results, computing the SOG requires expensive conversion of linguistic data into bit-level operators using logic synthesis tool Yosys \cite{wolf2013yosys}, and a Verilog-to-graph parser, which is outside the scope of this work.

Finally, some works take a step further and try to assist circuit design by annotating which parts of HDL is most critical to achieve quality-of-result metrics. For example, \cite{sengupta2023early} attempts to identify timing critical components based on path delay prediction. The AST of each Verilog design is extracted and converted into a graph, with nodes representing IO ports, registers or behavior logic. Behavioral paths are extracted from the graph and used for path-level feature generation. Delay labels of timing paths are generated using commercial synthesis tools, and are assigned to corresponding behavior paths with the same start and end points. By training an XGBoost model on the resulting features, the authors achieve an average classification accuracy of 91\%. Also similar is RTL-Timer~\cite{fang2024annotating}, which ensembles four bit-level circuit representations to predict the post-logic synthesis endpoint arrival time. Such predictions can then be mapped to registers in HDL code to identify critical code paths. Just as in the work of \cite{fang2023masterrtl}, however, these representations are bit-level rather than word-level and require pre-processing by logic synthesis tools like Yosys.

\subsection{LLMs for Verilog}

Large language models (LLMs) such as GPT~\cite{ouyang2022training} and Llama~\cite{touvron2023llama} have achieved exceptional success in various natural language tasks and have expanded their success to programming languages as well. Although excellent on generalist programming languages like Python or C++, these models have been trained on the relatively small amount of HDL code that is publicly available on the internet, and therefore have performed poorly on  Verilog benchmarks like VerilogEval~\cite{liu2023verilogeval} and RTLLM~\cite{lu2024rtllm}. This has motivated further work to build LLMs with a higher-degree of knowledge of hardware description languages. Both CodeGen-Verilog~\cite{thakur2023benchmarking} and VeriGen~\cite{thakur2024verigen} used a combination of customized Verilog datasets from code repository website GitHub\footnote{www.github.com} and various textbooks to fine-tune code LLMs. Finally, RTLCoder~\cite{liu2023rtlcoder} used the GPT 3.5 language model \cite{brown2020language} to generate further Verilog data, in a form of data augmentation, while CodeV ~\cite{zhao2024codev} used the same model to generate natural language description of real world Verilog code through multi-level summarization.

Besides Verilog code generation from natural language description, LLMs were also explored for other EDA-related tasks. RTLFixer~\cite{tsai2023rtlfixer} employed Retrieval-Augmented Generation (RAG) and ReAct prompting techniques to interactively debug syntax errors in Verilog code, and achieved remarkable improvement in success rates in the VerilogEval benchmark. ChipNemo~\cite{liu2023chipnemo} explored the application of LLMs in chip design process and adopted several domain adaptation techniques to train an LLM for various applications including assistant chatbots, EDA script generation, and bug summarization and analysis. Finally, ChatEDA~\cite{wu2024chateda} used code LLMs as an agent to autonomously complete the entire chip design flow from HDL code to the Graphic Data System Version II (GDSII) by managing task planning, script generation and task execution. We refer the reader to the extensive survey of \cite{zhong2023llm4eda} for more details on the application of LLMs in electronic design automation and future research directions in this field.

\subsection{Alignment of LLM and GNN Embeddings}

The multimodal alignment regularizer we propose during training also relates to the broader literature on tuning large language models to align with a pre-trained graph neural network, to incorporate its capabilities. 

The work closest to ours is that of \cite{mavromatis2023train}, who train a language model to perform a node classification task while adding a regularizer that encourages the predictive distributions to match a pre-trained graph neural network model. The language model makes predictions by passing the graph as input, and extracting the representation corresponding to a final [CLS] classification token. Also similar is \cite{zou2023pretraining}, which jointly trains a language model and a graph neural network on a common ``context graph prediction'' task which encourage alignment of their representations. They then discard the graph neural network and only keep the language model, so that topological characteristics best captured by graph convolutions can be said to have been incorporated in the language model.

More generally, there is a large literature on integrating pretrained graph neural networks with language models by training an adaptive module \cite{liu2024git,liu2023molca,chai2023graphllm,tang2024graphgpt,cao2023instructmol}, allowing the language model to receive inputs from the graph neural network. Alternatively, multiple works have interlaced graph neural network layers and language model layers \cite{yasunaga2021qa,zhang2022greaselm,jin2023patton}.  In either case, some kind of training is necessary to allow for interactions between the graph neural network and the language model, although the result is not distillation of the graph neural network's perspective into the language model per se.

\section{Methodology} \label{sec:method}

We now present our VeriDistill approach in detail. As described in the introduction, turning a high-level description of a circuit in a Hardware Description Language like Verilog into a physical description ready for manufacturing is a computationally expensive process involving several steps, each with an associated intermediate representation describing progressively lower-level elements of the circuit. Our goal is to predict low-level quality-of-result metrics, like area and delay, from the high-level representation, namely the HDL code to allow for fast iterative improvement of the RTL design. Figure \ref{fig:overview} provides and overview of logic synthesis as well as the training and inference pipelines.

\subsection{Model}

Our model takes as input Verilog code, which is fed to a Large Language Model (LLM). This LLM has been specifically fine-tuned on Verilog code generation. The code is first split into a sequence of tokens, which are then fed in parallel in the LLM. As an output, the LLM produces a sequence of high-dimensional ``hidden state'' vectors, one for each token that is inputted to the LLM. We average these hidden states, producing a single vector. This vector is then fed to a feedforward neural network, composed of several linear layers with nonlinear activations, which finally outputs the QoR estimate.


\subsection{Training}

We produce a training set of circuits with Verilog code for which the expensive logic synthesis process has been performed, so we know their QoR metric (such as area or delay). 
In addition, as an intermediate product of the logic synthesis process, an LUT graph is produced immediately following the logic optimization phase, which we save. This yields a collection of training triples $\mathcal{D} = \{(X_\text{Verilog}, X_\text{LUT}, y_\text{QoR})\}$.

\paragraph{Supervised learning:} Given such a dataset, we treat our problem by supervised machine learning. The LLM, which has been pretrained on Verilog code, is kept frozen, so that only the FNN gets updated. In a training step, the Verilog code $X_\text{Verilog}$ is fed to the VeriDistill model to produce a prediction $\hat{y}_\text{QoR}$. This prediction is compared in mean-squared error loss with the true QoR metric $y_\text{QoR}$ as a supervised learning loss
\begin{align}\label{eq:sl-loss}
\mathcal{L}_\text{SL} = \big(\hat{y}_\text{QoR}-y_\text{QoR}\big)^2.
\end{align}

\paragraph{Low-level knowledge distillation:} In practice, training only with the supervised learning loss leads to limited performance. One potential explanation is that there is too much of a gap between a high-level circuit description like Verilog and the low-level metrics we purport to predict. Intuitively, to perform high-quality predictions, we would want the model to possess some degree of understanding of lower-level circuit design while still only taking Verilog code as input.

We propose the following approach to address this problem. Prior to training, we pretrain a Graph Neural Network (GNN) to predict the same QoR metric as VeriDistill, but from the Look-Up-Table (LUT) graph $X_\text{LUT}$ of the circuit obtained after optimization using Yosys~\cite{wolf2013yosys}. This graph, which can be seen as an alternative to the more popular And-Inverter Graph (AIG) format, is particularly suitable for GNN training as it is compact with rich node information. Moreover, as a circuit representation, it sits intermediate between a high-level description of the circuit encoded in the Verilog code, and a physical circuit description. Prediction from LUT graphs is thus easier than prediction from Verilog code, but not completely trivial either.

The GNN architecture we adopt is composed of a sequence of graph convolutions, followed by joint mean and max pooling, and a sequence of linear layers. We pretrain it using the supervised learning loss (\ref{eq:sl-loss}) until good predictive performance is achieved. Then, during the VeriDistill training, we keep the GNN weights frozen and we propose to encourage the last-layer activations of the VeriDistill model $z^{(-1)}_\text{VeriDistill}$ to resemble those of the GNN model $z^{(-1)}_\text{GNN}$, despite these models operating on different inputs. We perform this simply by adding a mean-square error loss
\begin{align}\label{eq:kd-loss}
\mathcal{L}_\text{KD} = \big\|z^{(-1)}_\text{VeriDistill}-z^{(-1)}_\text{GNN}\big\|_2^2
\end{align}
in the total loss. As the weights of the GNN are pretrained and kept frozen while the VeriDistill model is being trained, this is effectively a form of knowledge distillation from the GNN to the VeriDistill model.

\begin{figure}[t]
    \centering
    \includegraphics[clip, trim={20pt 10pt 70pt 10pt}, width=0.95\linewidth]{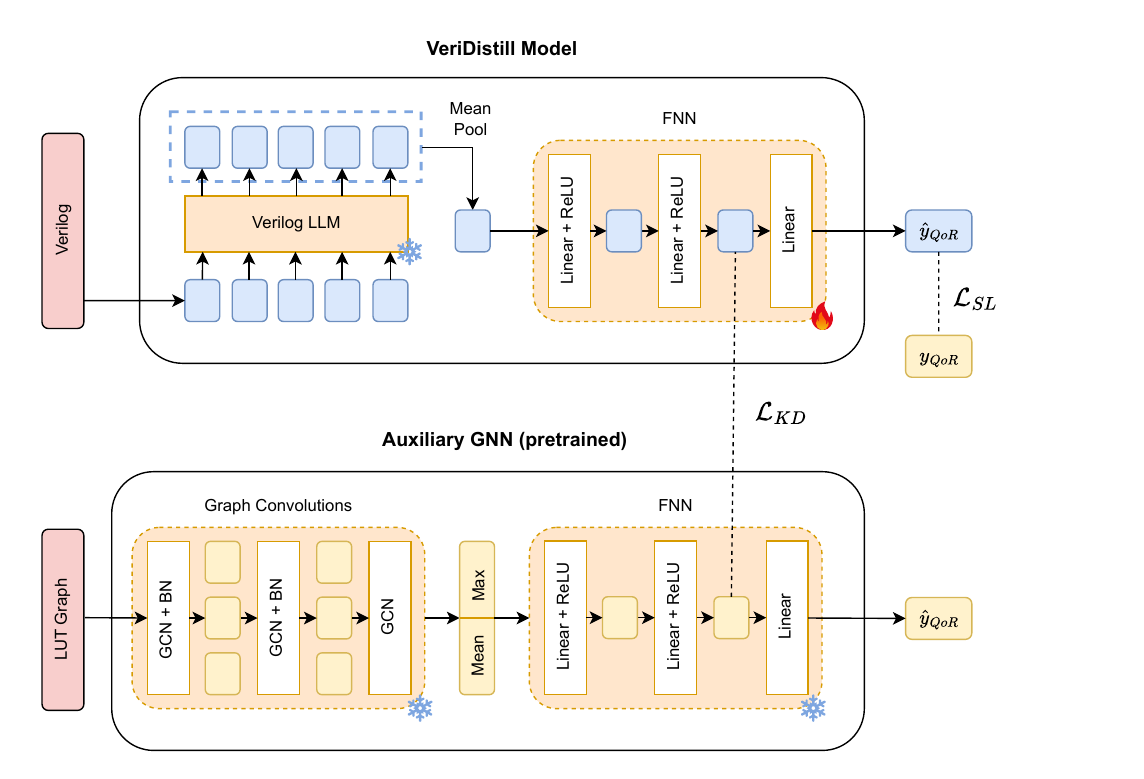}
    \caption{The training procedure. The Verilog training examples are passed to the VeriDistill model, which produces predictions of the QoR metric. These predictions are scored against the true QoR values by a mean-squared error supervised learning loss. In addition, the LUT graph representation resulting from logic optimization is fed to an auxiliary GNN model, pretrained to perform the same QoR prediction task. The hidden representations at the last layer of both the VeriDistill and GNN models is extracted, and a mean-square error knowledge distillation loss encourages these two representations to be similar, despite having different inputs. Both the pretrained GNN and LLMs modules are kept frozen during training.}
    \label{fig:training}
\end{figure}

\paragraph{Total loss:} We balance the importance given to the knowledge distillation compared to the supervised learning objective using a hyperparameter factor $\alpha$, yielding the final loss
\begin{align*}
\mathcal{L} = \alpha\mathcal{L}_\text{SL} + (1-\alpha)\mathcal{L}_\text{KD}.
\end{align*}
A diagram describing the VeriDistill training process is provided as Figure \ref{fig:training}.
\section{Experiments} \label{sec:experiments}

This section is organized as follows: We begin by presenting the implementation details of our experimental setup in Section \ref{sec:exp_setup}, including hardware, model, and training hyperparameters. Next, we describe the dataset used and the data preprocessing steps for training and evaluation in Section \ref{sec:dataset}. We then introduce the baseline methods and their implementation details in Section \ref{sec:baselines}. Finally, we present the results on the main datasets and a study on unseen out-of-distribution circuits in Sections \ref{sec:results} and \ref{sec:openABCD}.

\subsection{Experimental Setup} \label{sec:exp_setup}

\begin{table*}[h]
\centering
\begin{tabularx}{0.84\linewidth}{lrrrrrrrrr@{}}
\toprule
\multirow{2}{0.5cm}{method} & \multicolumn{4}{c}{Area} & \multicolumn{4}{c}{Delay} \\ 
\cmidrule(lr){2-5}\cmidrule(lr){6-9}
& MAE $\downarrow$ & R2 $\uparrow$ & MAPE $\downarrow$ & RSE $\downarrow$ & MAE $\downarrow$ &  R2 $\uparrow$ & MAPE $\downarrow$ & RSE $\downarrow$ \\ \midrule 
LUT-GNN (Teacher)\hspace{-7pt} & 0.251 & 0.955 & 0.309 & 0.045 & 0.109 & 0.948 & 0.023 & 0.052 &  \\ \hline
\rule{0pt}{2.3ex}AST-XGBoost & 1.497 & 0.255 & 71.205 & 2.899 & 0.480 & 0.280 & 0.108 & 2.564 &\\
AST-GNN &  0.893 & 0.661 & 1.435 & 0.339 & 0.317 & 0.604 & 0.071 & 0.396 \\
AST-GNN w/ KD & 0.892 & 0.666 & 1.647 & 0.334 & 0.315 & 0.619 & 0.071 & 0.381 &  \\ \hline
DeepSeek + Decoder & 1.119 & 0.548 & 2.004 & 0.452 & 0.401 & 0.478 & 0.094 & 0.522 & \\ 
CodeV + Decoder & 0.991 & 0.629 & 1.69 & 0.371 & 0.367 & 0.533 & 0.086 & 0.467 & \\
VeriDistill (DeepSeek) & 0.497 & 0.867 & 0.863 & 0.133 & 0.23 & 0.793 & 0.053 & 0.207 &  \\
VeriDistill (CodeV) & 0.482 & 0.872 & 0.784 & 0.128 & 0.236 & 0.781 & 0.054 & 0.219 &  \\
\bottomrule
\end{tabularx}
\caption{\centering The performance of different Verilog models on the test dataset, where the best result for each metric is bolded. In addition, we report the performance of the teacher model trained on the LUT graphs, which serves as an upper-bound.}
\label{tab:results-optsetting}
\end{table*}

We use the following implementation of the model. We employ DeepSeek-Coder-V2-Lite-Base \cite{deepseekai2024} and CodeV-7B (CodeLlama) \cite{zhao2024codev} as the Verilog LLM, and three layers with ReLU activations in the feedforward neural network. The model takes as input strings, which are broken into a sequence of tokens in the LLM's vocabulary. The language model processes these inputs into a sequence of the same length, made up of 512-dimensional vectors. After mean pooling, the resulting vector is passed to the feedforward neural network, which uses 512-dimensional activations, before making the final prediction. In particular, this architecture means that the last-layer activations $z^{(-1)}_\text{VeriDistill}$ are 512-dimensional. Results for other variants of CodeV-7B \cite{zhao2024codev} as the Verilog LLM are included in Appendix B.\\

The auxiliary GNN teacher model takes a LUT graph with 16-dimensional node attributes, and passes it through three 64-dimensional graph convolutional layers interleaved with batch normalization layers. After concatenation of the mean and max pooling outputs, the 128-dimensional vector is passed through three 512-dimensional linear layers with ReLU activations before the final prediction. Thus, in particular, the last-layer activations $z^{(-1)}_\text{GNN}$ are 512-dimensional, matching with those of the VeriDistill model.

We implement VeriDistill and the baselines using the \texttt{PyTorch} and \texttt{PyG} libraries. Models which do not use our knowledge distillation procedure are trained using the ReduceLROnPlateau scheduler with initial learning rate 1e-3, patience set to 30 epochs and factor set to 0.5. In contrast, models involving our knowledge distillation procedure are trained using the CosineAnnealingLR \cite{DBLP:conf/iclr/LoshchilovH17} scheduler, with an initial learning rate of 1e-3 and number of iterations set to 50. We start the training process with $\alpha = 0.5$, and increase $\alpha$ to $0.75$ and $1$ at epochs $150$ and $250$. The idea is put less emphasis on knowledge distillation at every warm re-start. We find that this approach results in marginal gain compared to other optimization methods. 

All models are trained until full convergence. Since the LLM is kept frozen during training, it was possible to save training time by extracting the forward pass through the LLM only once and saving it. We performed this phase on a machine with 8 Nvidia V100 GPUs with 32GB of memory and 32 Intel(R) Xeon(R) Gold 6140 CPUs. Once the hidden state a then trained each model following the procedure detailed in the paper on the same machine using a single V100 GPU with 1024 minibatch sizes. The training times for each model are summarized in Appendix F.

\subsection{Datasets} \label{sec:dataset}
We train and evaluate on two separate datasets. The first dataset is used for training, validation, and testing of all the methods, while OpenABCD contains out-of-distribution circuits aiming to challenge VeriDistill and determine its ability to generalize.

\paragraph{Customized Dataset} To train and evaluate our proposed method, we collect 18.4k Verilog examples provided by \cite{pei2024betterv} and 5.8k from \cite{Thakur2022BenchmarkingLL}. These Verilog examples are obtained from open-source GitHub repositories and textbooks and have been verified for syntax correctness. We use an open-sourced EDA platform OpenROAD~\cite{ajayi2019openroad} with 7nm technology PDK provided to conduct logic synthesis and record post-synthesis labels of area and delay. We convert the AIG graphs obtained after logic optimization into LUT graphs and save them for training the auxiliary GNN model. 

Note that a substantial fraction of the code snippets end up being functionally incorrect and failing some stage of the logic synthesis pipeline. Since we require functionally correct examples for their QoR metric to be well-defined, we removed such examples during the preprocessing. In addition, although not strictly a problem for our method, one of the competing baselines requires extracting the Abstract Syntax Tree (AST) of the Verilog, which is obtained by running a parser on the code. The parser was unable to produce AST representations for a small fraction of the instances (FRACTION\%), which we removed from consideration. The resulting dataset, after filtering bad examples, ended up having 16k examples, which we split into training, validation, and test sets with a ratio of 0.75/0.1/, respectively.  Details about the dataset and label distributions can be found in Appendix A.

We note that OpenROAD provides two optimization recipes for the logic synthesis process: ``ABC\_AREA=1" for area optimization and ``ABC\_SPEED=1" for timing optimization. The results reported under Section \ref{sec:results} are produced under the speed optimization. We report the results under area optimization in Appendix C. We find that our approach works as well under different recipe optimization settings.

\paragraph{OpenABCD} Additionally, we consider data provided by~\cite{chowdhury2021openabc} to evaluate the transferability of our method to unseen circuits. The OpenABCD dataset consists of functionally diverse designs such as bus communication protocols, computing processors, digital signal processing cores, cryptographic accelerators and system controllers.

\begin{figure*}[h]
    \centering
    \includegraphics[width=0.18\linewidth]{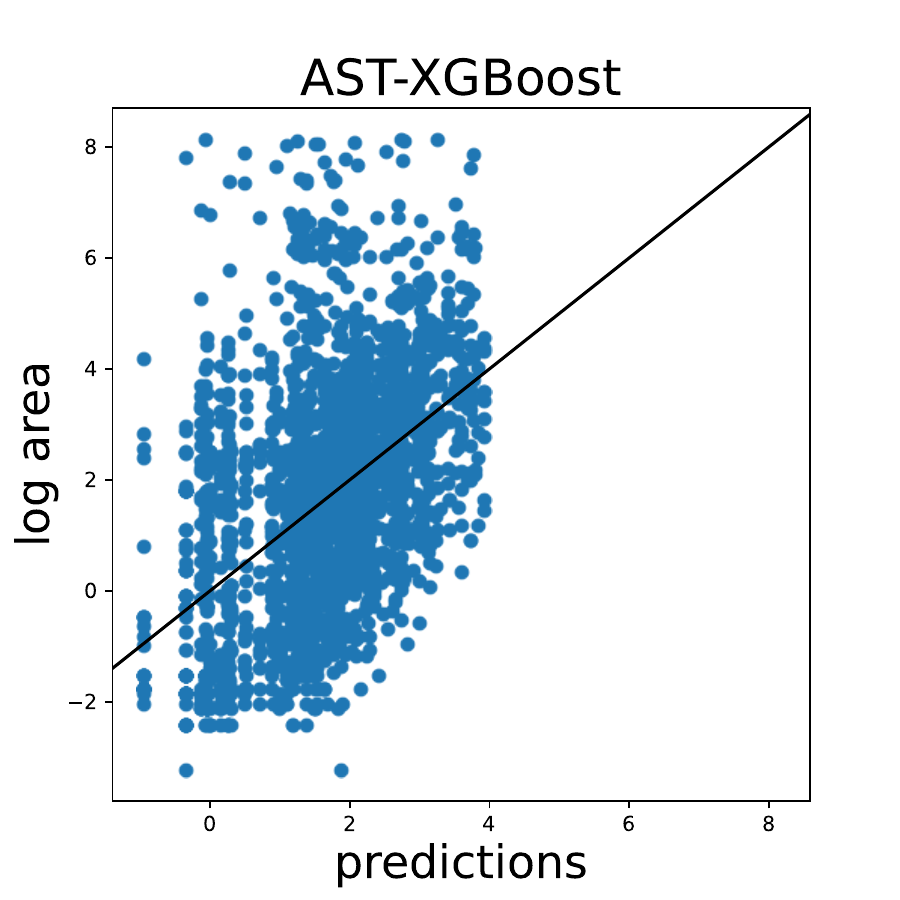}
    \includegraphics[width=0.18\linewidth]{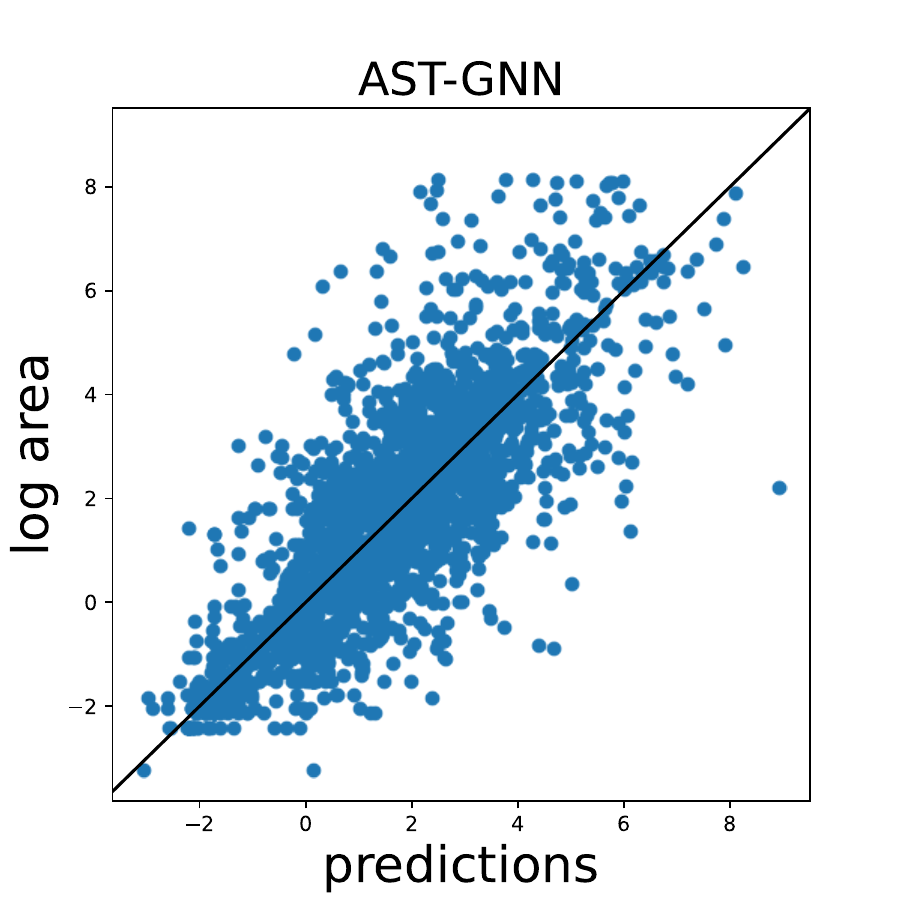}
    \includegraphics[width=0.18\linewidth]{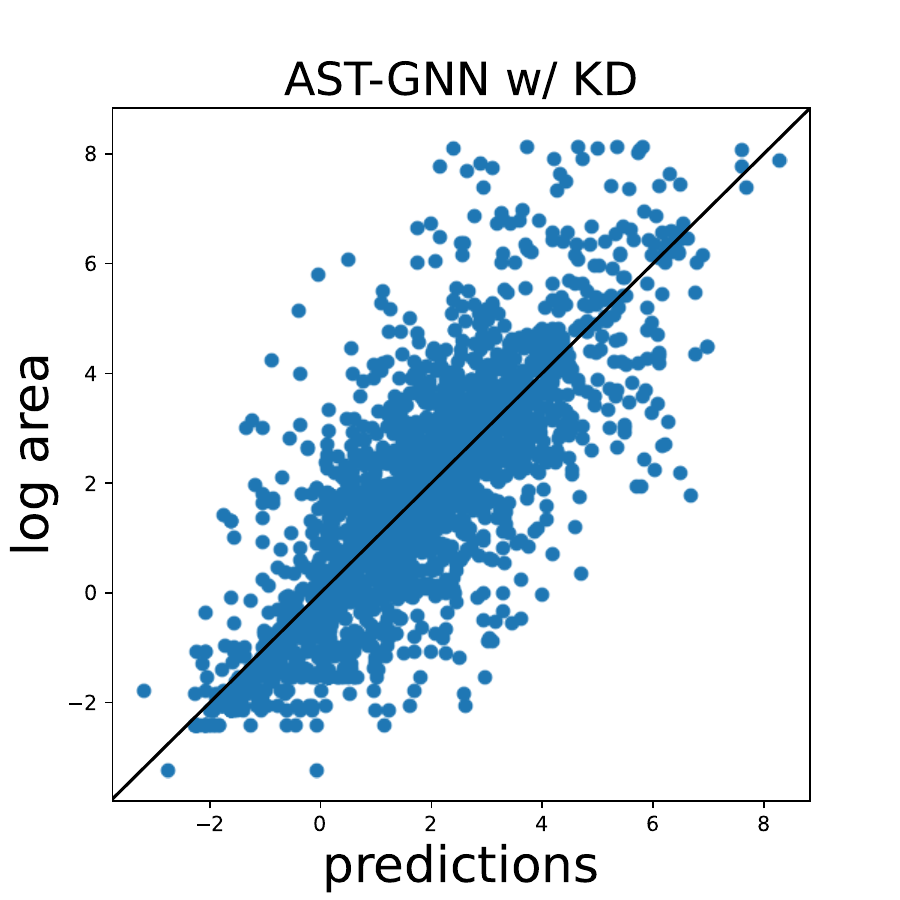}
    \includegraphics[width=0.18\linewidth]{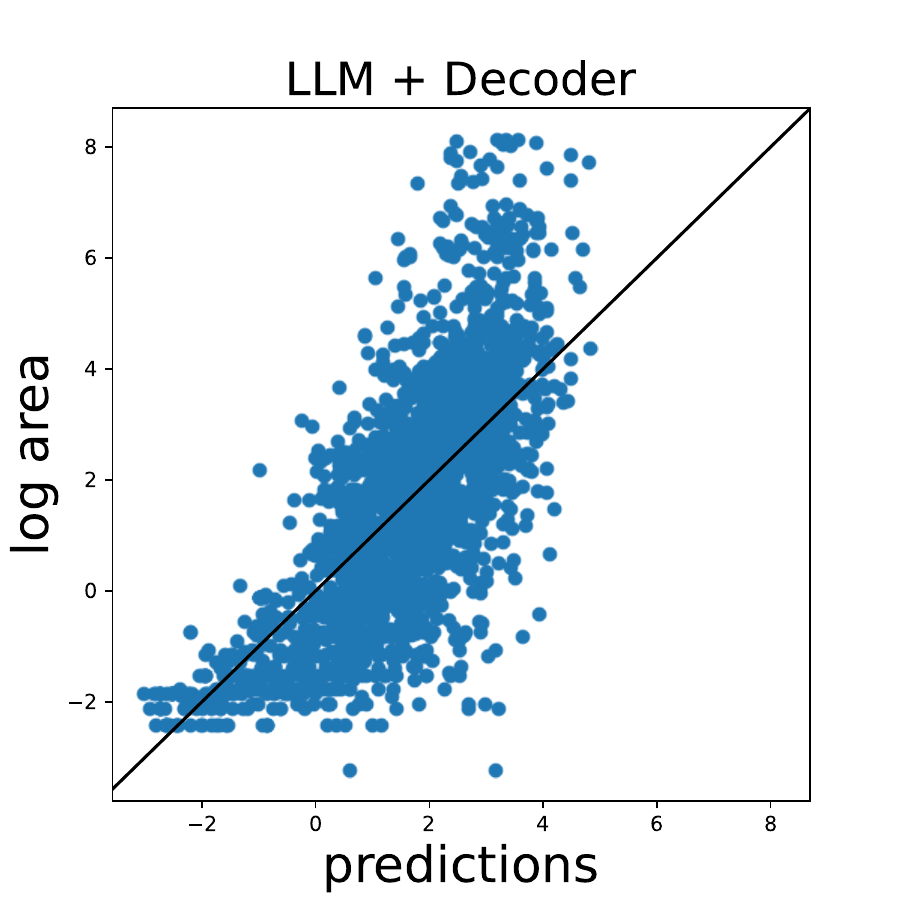}
    \includegraphics[width=0.18\linewidth]{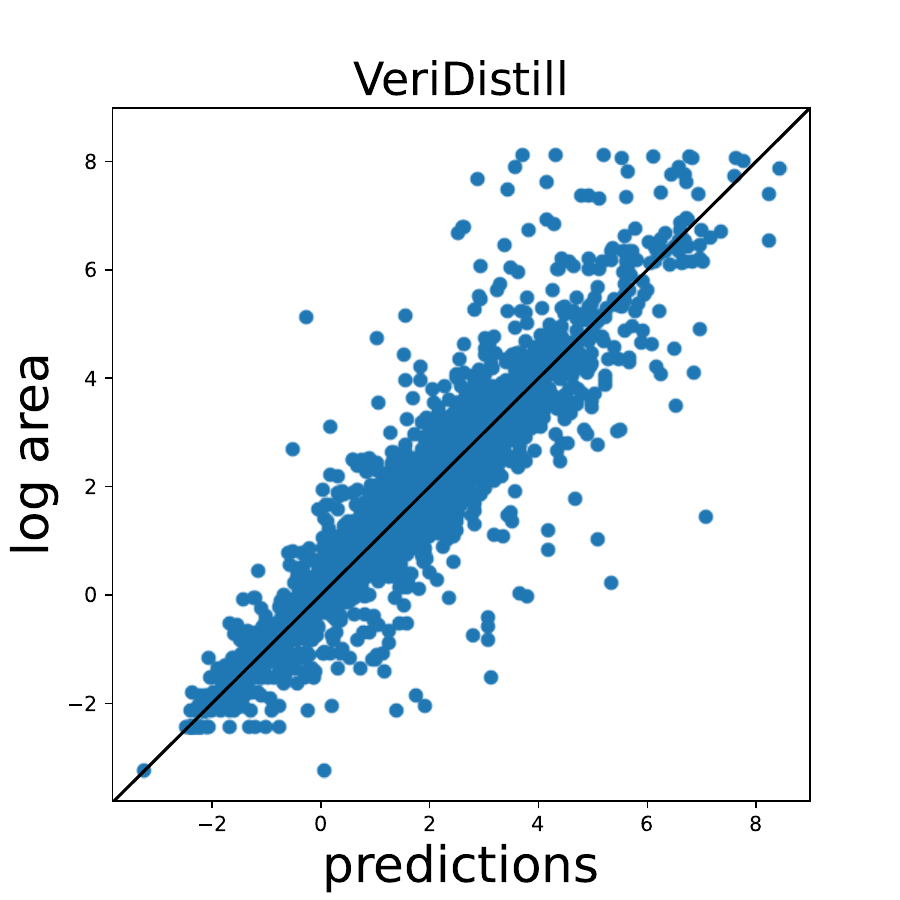}
    \includegraphics[width=0.18\linewidth]{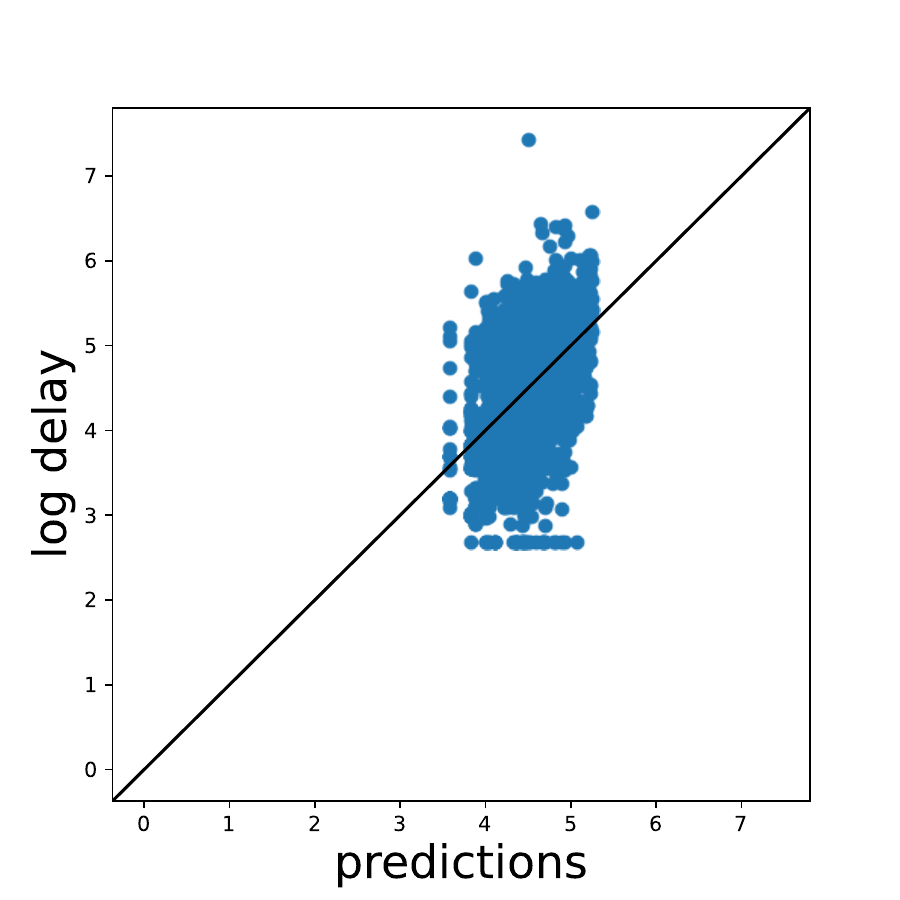}
    \includegraphics[width=0.18\linewidth]{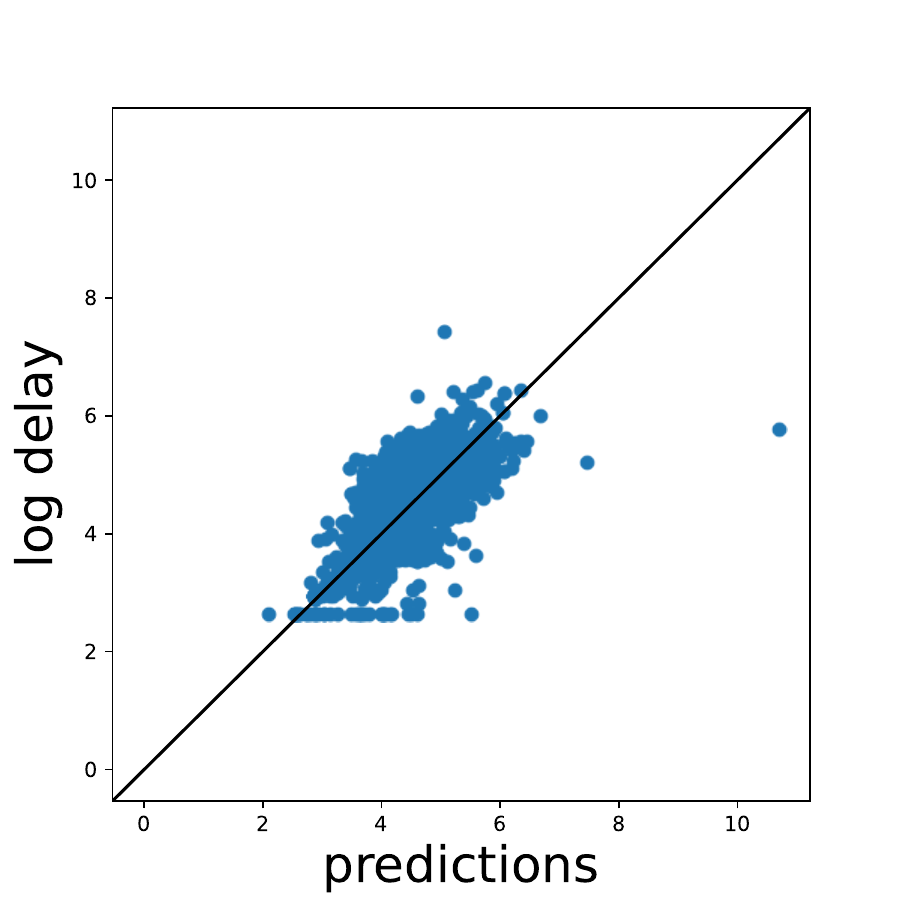}
    \includegraphics[width=0.18\linewidth]{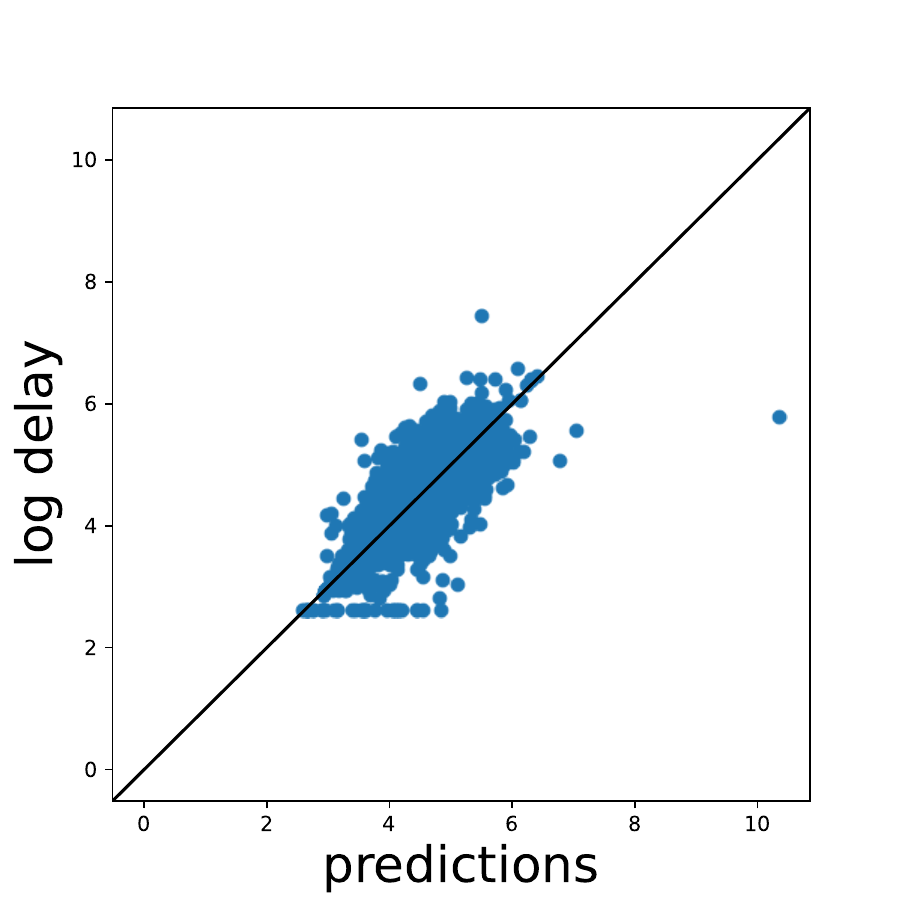}
    \includegraphics[width=0.18\linewidth]{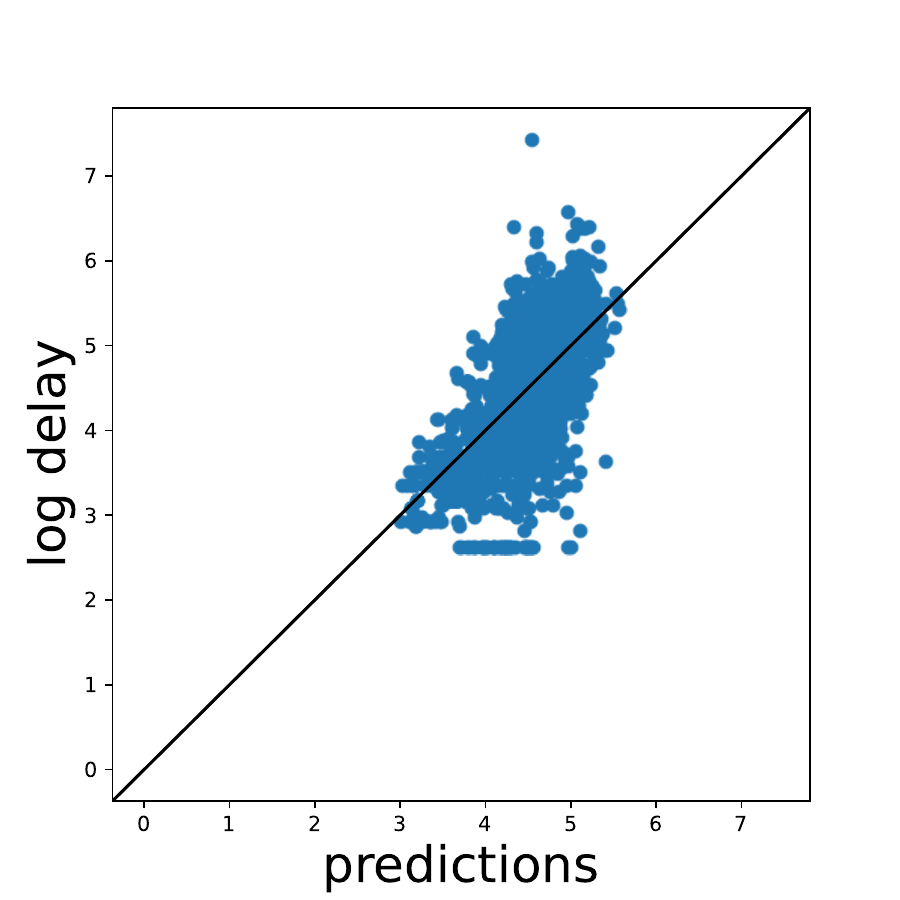}
    \includegraphics[width=0.18\linewidth]{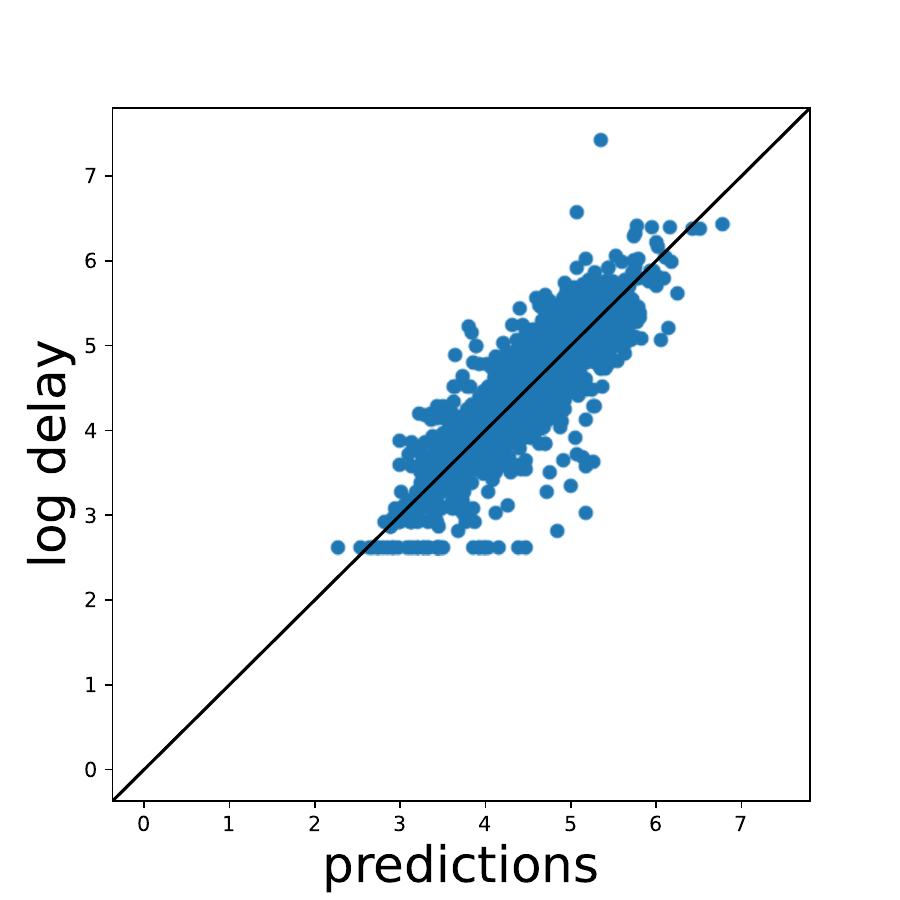}
    \caption{\centering Prediction vs. target on test data, where DeepSeek-V2-Lite is utilized as the LLM. The predicted values using different methods are plotted against the targets. (Top) Area prediction. (Bottom) Delay prediction.}
    \label{fig:scatter}
\end{figure*}

\subsection{Baselines} \label{sec:baselines}
While many prior works have attempted to predict post-synthesis circuit quality at the RTL-stage, none of them perform prediction directly from source Verilog files. Several works rely on lower-level circuit representation that requires extra processing using logic synthesis tools \cite{zhou2019primal,fang2023masterrtl}. Utilizing low-level circuit representations as input is advantageous for the circuit quality prediction. However, in some cases, obtaining the low-level description can be prohibitively expensive. In addition, obtaining the low-level description relies on external processing tools which are prone to errors. As such, we compare our method to approaches which take the raw Verilog or the AST representation of the circuit. 

We adopt the method proposed by \cite{sengupta2022iccad} as our baseline. It relies on AST representations that can be easily converted from Verilog source files. We implement the method based on description in \cite{sengupta2022iccad}. Verilator~\cite{snyder2004verilator} is used to convert each source Verilog into its respective AST representation, which can be represented as a graph. The nodes in the graph represent one of the following five semantic categories from the source Verilog (\texttt{root}, \texttt{variable}, \texttt{operation}, \texttt{constant}, \texttt{edge}), while edges are created between nodes with connections. We implement three variants of the AST-based method:

\paragraph{AST-XGBoost} We compute the following features: $(i)$ the total number of input bits, $(ii)$ the total number of output bits, $(iii)$ the longest path in the AST, $(iv)$ the frequency of each node type in the graph and $(v)$ the frequency of each logic type in the graph. The features are concatenated to form a feature vector with 108 features \footnote{ 108 = 1 + 1 + 1 + 5 + 100 features coming from feature categories $(i) ...., (v)$}. We perform a thorough hyper-parameter selection using grid search and employ early stopping to prevent over-fitting. 

\paragraph{AST-GNN w/o KD} The AST-GNN model takes in the following features per node: $(i)$ the total number of input bits, $(ii)$ the total number of output bits, $(iii)$ the node semantic type and $(iv)$ the node operation type. Each feature is represented via a one-hot vector and is projected to a 4-dimensional space via a linear layer. The final node features consist of a ($4 \times 4$) = 16-dimensional vector. We cap the number of input/output bits to 200, since 99.9 percent of the nodes in the dataset have less than 200 input/outputs. The AST-GNN model utilizes the same hyperparameters and architecture as the auxiliary GNN model used for the knowledge distillation objective in VeriDistill.

\paragraph{AST-GNN w/ KD} We propose a third baseline, where the AST-GNN model is guided by the LUT GNN model. The baseline utilizes the same student-teacher knowledge distillation as our method. We introduce this baseline to demonstrate the effectiveness of utilizing an LLM in the student network. 

\subsection{Main results} \label{sec:results}
We first summarize the results of our main experiment, where we train and test the model on the large Customized Dataset (see Section \ref{sec:dataset}). Table \ref{tab:results-optsetting} outlines the performance of different models on the test set. As can be seen, our proposed method, utilizing both the LLM as an encoder and knowledge distillation, outperforms other baselines across all the metrics, especially with area prediction. Interestingly, simply using a decoder on the LLM representation performs worse than the previous state-of-the-art, while knowledge distillation on the AST-GNN model has almost no effect. Only when both are used together is there profound impact on performance, which suggests our knowledge distillation procedure is crucial in fully exploiting the richness of the LLM representations. 

The CodeV + Decoder model greatly outperforms the DeepSeek-V2-Lite + Decoder model. This is due to the fact that CodeV representations are more aligned with Verilog semantics, since the model is specially fine-tuned on Verilog data after being trained on various programming languages. However, the performance gap between the two VeriDistill models  is much smaller. This results hints that using knowledge distillation can, to a high degree, mitigate the lack of additional fine-tuning of the base LLM on Verilog. Hence, our method can potentially be easily applied on top of other code LLMs without the need for further fine-tuning the base LLM on Verilog data.

\vspace{-0.5em}

\begin{table}[H]
\small
\begin{tabular}{l|ccc|cc}
\toprule
\texttt{IP} & \texttt{IO} & \texttt{Nodes} & \texttt{Edges}  & \texttt{Lines} & \texttt{Tokens} \\ 
\midrule
aes & 1212 & 28925 & 58379  &  1406 & 21305 \\
aes\_secworks & 5691 & 40778 & 84160 & 2443  & 28630\\
aes\_xcrypt & 3780 & 45840 & 93485 & 985  & 16308\\
des3\_area & 367 & 4971 & 10006 & 2545 & 44650\\
dft & 75014 & 245046 & 527509 & 4637  & 59292\\
dynamic\_node & 5283 & 18094 & 38763  & 6251  & 63011\\
ethernet & 21153 & 67164 & 144750 & 10841  & 131015\\
fir & 761 & 4558 & 9467 &  307 & 2892\\
fpu & 1041 & 29623 & 59655 & 1910  & 27060\\
i2c & 305 & 1169 & 2466 &  1246 & 13690\\
idft & 75022 & 241552 & 520523  & 4638  & 59356\\
iir & 935 & 6978 & 14397 & 395  & 3870\\
mem\_ctrl & 2149 & 16307 & 37146 &  5880 & 70632\\
pci & 6586 & 19547 & 42251 & 22692  & 306937\\
sasc & 260 & 613 & 1351 & 597  & 5783\\
sha256 & 2985 & 15816 & 32647 & 1054  & 10551\\
simple\_spi & 296 & 930 & 1992 &  463 & 5010\\
spi & 492 & 4219 & 8676  &  794 & 10348\\
ss\_pcm & 194 & 462 & 896 & 223  & 2173\\
tv80 & 997 & 11328 & 23017 & 4736  & 56461\\
usb\_phy & 222 & 487 & 1064 & 1102  & 10317\\
vga\_lcd & 34385 & 105334 & 227731 & 5078  & 54555\\
wb\_conmax & 4197 & 47840 & 97755 & 7108  & 108718\\
\bottomrule
\end{tabular}
\caption{OpenABCD  circuit statistics. \texttt{IO}, \texttt{Node} and \texttt{Edges} are the number of primary inputs/outputs, AIG nodes and AIG edges of the circuits. \texttt{Lines} and \texttt{Tokens} refer to the number of lines and tokens in the Verilog RTL file.}
\label{tab:openabcd}
\end{table}

\vspace{-1.0em}

\begin{table*}[t]
\centering
\begin{tabular}{l|c|cc|c|cc}
\multicolumn{1}{c}{}   & \multicolumn{3}{c}{Area}                                                                  & \multicolumn{3}{c}{Delay}                                                                  \\ 
\hline
\multicolumn{1}{c}{IP} & \multicolumn{1}{c}{GT} & \multicolumn{1}{c}{AE (wo/ KD)} & \multicolumn{1}{c}{AE (w/ KD)} & \multicolumn{1}{c}{GT} & \multicolumn{1}{c}{AE (wo/ KD)} & \multicolumn{1}{c}{AE (w/ KD)}  \\ 
\hline
aes                                                                               & 7.3                                                                         & 2.909          & \textbf{2.746}                                                               & 5.886                                                                        & 0.442          & \textbf{0.212}                                                             \\
aes\_secworks                                                                     & 7.629                                                                       & 3.763          & \textbf{0.631}                                                               & 6.33                                                                         & 1.089          & \textbf{0.494}                                                             \\
aes\_xcrypt                                                                       & 8.013                                                                       & 3.742          & \textbf{3.319}                                                               & 6.363                                                                        & 0.949          & \textbf{0.713}                                                             \\
des3\_area                                                                        & 5.772                                                                       & 1.971          & \textbf{1.068}                                                               & 6.05                                                                         & 0.639          & \textbf{0.22}                                                              \\
dft                                                                               & 9.671                                                                       & 5.813          & \textbf{4.004}                                                               & 5.557                                                                        & 0.267          & \textbf{0.167}                                                             \\
dynamic\_node                                                                     & 7.046                                                                       & \textbf{3.379} & 4.965                                                                        & 5.986                                                                        & \textbf{0.937} & 1.245                                                                      \\
ethernet                                                                          & 8.464                                                                       & 5.134          & \textbf{4.395}                                                               & 5.883                                                                        & 0.801          & \textbf{0.525}                                                             \\
fir                                                                               & 5.04                                                                        & 2.12           & \textbf{0.829}                                                               & 5.765                                                                        & 0.83           & \textbf{0.527}                                                             \\
fpu                                                                               & 7.092                                                                       & 4.077          & \textbf{0.977}                                                               & 7.714                                                                        & 2.682          & \textbf{1.574}                                                             \\
i2c                                                                               & 4.193                                                                       & 2.038          & \textbf{1.006}                                                               & 5.412                                                                        & 0.692          & \textbf{0.157}                                                             \\
idft                                                                              & 9.668                                                                       & 5.731          & \textbf{4.002}                                                               & 5.537                                                                        & 0.208          & \textbf{0.14}                                                              \\
iir                                                                               & 5.508                                                                       & 2.297          & \textbf{0.983}                                                               & 5.768                                                                        & 0.736          & \textbf{0.572}                                                             \\
mem\_ctrl                                                                         & 6.366                                                                       & 2.597          & \textbf{0.748}                                                               & 6.155                                                                        & 1.067          & \textbf{0.585}                                                             \\
pci                                                                               & 7.124                                                                       & 3.179          & \textbf{3.162}                                                               & 5.765                                                                        & \textbf{0.574} & 0.594                                                                      \\
sasc                                                                              & 3.795                                                                       & 0.971          & \textbf{0.583}                                                               & 5.17                                                                         & \textbf{0.229} & 0.427                                                                      \\
sha256                                                                            & 6.594                                                                       & 2.259          & \textbf{0.043}                                                               & 5.587                                                                        & 0.305          & \textbf{0.194}                                                             \\
simple\_spi                                                                       & 4.07                                                                        & 2.151          & \textbf{0.985}                                                               & 5.263                                                                        & 0.636          & \textbf{0.173}                                                             \\
spi                                                                               & 5.363                                                                       & 2.321          & \textbf{1.968}                                                               & 5.82                                                                         & 0.896          & \textbf{0.734}                                                             \\
ss\_pcm                                                                           & 3.309                                                                       & 0.78           & \textbf{0.192}                                                               & 4.89                                                                         & \textbf{0.106} & 0.234                                                                      \\
tv80                                                                              & 6.235                                                                       & 3.126          & \textbf{1.259}                                                               & 6.178                                                                        & 1.097          & \textbf{0.793}                                                             \\
usb\_phy                                                                          & 3.412                                                                       & 0.359          & \textbf{0.172}                                                               & 4.92                                                                         & \textbf{0.05}  & 0.14                                                                       \\
vga\_lcd                                                                          & 8.883                                                                       & 5.7            & \textbf{4.984}                                                               & 5.778                                                                        & 0.811          & \textbf{0.724}                                                             \\
wb\_conmax                                                                        & 7.689                                                                       & \textbf{3.253} & 4.665                                                                        & 5.866                                                                        & \textbf{0.832} & 1.521                                                                      \\ 
\hline
\multicolumn{1}{>{\centering\hspace{0pt}}m{0.15\linewidth}}{\textbf{Mean Value}} & \multicolumn{1}{>{\centering\hspace{0pt}}m{0.1\linewidth}}{}              & 3.029 & \multicolumn{1}{>{\centering\hspace{0pt}}m{0.11\linewidth}}{\textbf{2.073}} & \multicolumn{1}{>{\centering\hspace{0pt}}m{0.1\linewidth}}{}               & 0.734 & \textbf{0.551}                                                             \\
\hline
\end{tabular}
\caption{\centering The Absolute Error (AE) on OpenABCD circuits. VeriDistill with or without KD have been trained on customized datasets and used to predict post-synthesis area and delay of OpenABCD circuits without any finetuning. Due to the large length of the Verilogs, we utilize DeepSeek-V2-Lite with a context window of 128k tokens.}
\end{table*}

We gain further insight on the benefits of our approach by analyzing scatter plots of the predictions against the targets. As can be seen in Figure \ref{fig:scatter}, the baseline models perform well primarily on circuits with small delay and area but struggle with larger circuits, likely due to their lower representation in the training set.  In contrast, our model achieves consistently strong performance across circuits of all sizes. This contrast is particularly pronounced when comparing against the same model without knowledge distillation (LLM+Decoder), which indicates that our knowledge distillation procedure is crucial in allowing our model to perform well across the whole range of circuit sizes. The impact of knowledge distillation on the AST model is relatively minimal compared to its effect on the LLM-based model. This can be attributed to the enhanced alignment between the teacher representations and the LLM when used as the encoder. A visualization of the t-SNE projection of the final hidden space representations is provided in Appendix D to verify the above claim.

\subsection{Additional Out-of-Distribution Results} \label{sec:openABCD}
Finally, we evaluate how our knowledge-distillation procedure can impact the ability of the trained model to generalize to new out-of-distribution circuits. For this, we take our model, trained with and without knowledge distillation on our Customized Dataset, and apply it to instances in the OpenABCD benchmark (see Section~\ref{sec:dataset}). We outline the OpenABCD  circuit statistics in the Table \ref{tab:openabcd}. Due to the scarcity of large circuit Verilog data, the circuits from the OpenABCD benchmark are larger than the majority of the circuits present in our dataset (that is, the Verilog files contains more lines and the circuits have generally larger area and delay). 
As can be seen in Table \ref{tab:openabcd}, our knowledge distillation procedure systematically improves the LLM-based model's ability to transfer prediction performance on out-of-distribution instances, which differ significantly from those seen during training. In Appendix E, we compare the performance of VeriDistill with the AST based approaches on the OpenABCD benchmark, where VeriDistill outperforms the baselines on the well majority of the circuits.

\section{Conclusion} \label{sec:conclusion}
In this work, we propose a novel procedure to predict quality-of-result electronic circuit metrics from Verilog code, by training a small neural network model on Verilog LLM representations with a knowledge distillation regularizer which align its internal activations with those of a low-level GNN model. We show that this new model, which we call VeriDistill, outperforms previous approaches in prediction accuracy.

Beyond the potential of our method for future practical applications, our results underscore the value of the information encoded in the LLM’s representations for predicting circuit quality. Additionally, they highlight the crucial role of our knowledge distillation procedure in enabling downstream models to effectively leverage this information.

\bibliographystyle{named}
\bibliography{ijcai25}

\onecolumn
\appendix
\section*{Appendix}\label{sec:appendix}

\subsection*{A: Dataset Statistics}
\label{appendixG}
\begin{figure}[h]
     \centering
     \includegraphics[width=0.30\linewidth]{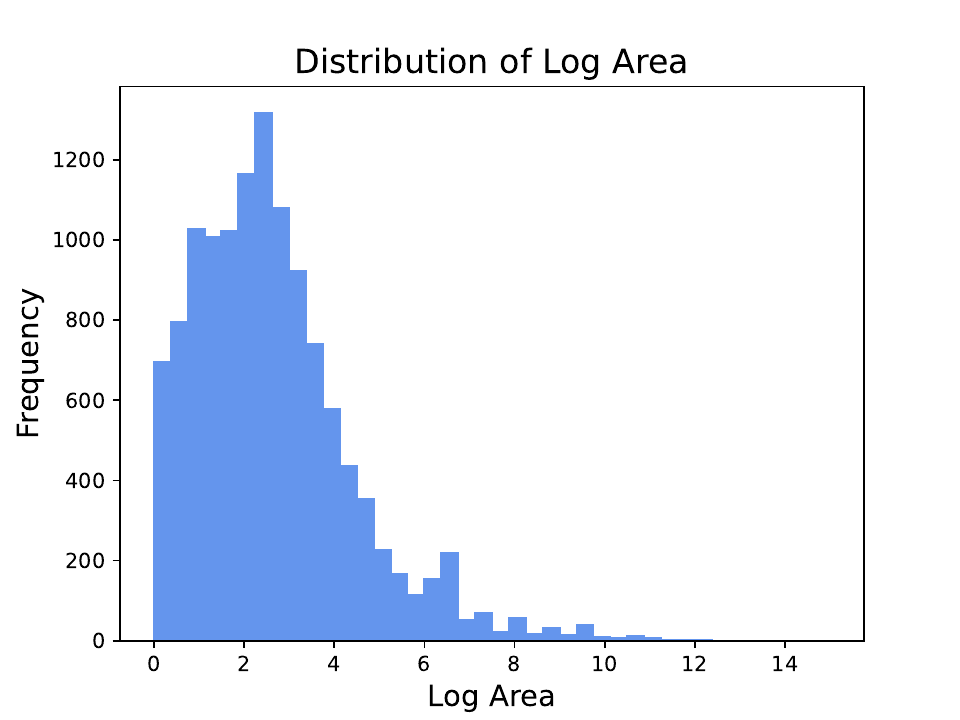}
     \includegraphics[width=0.30\linewidth]{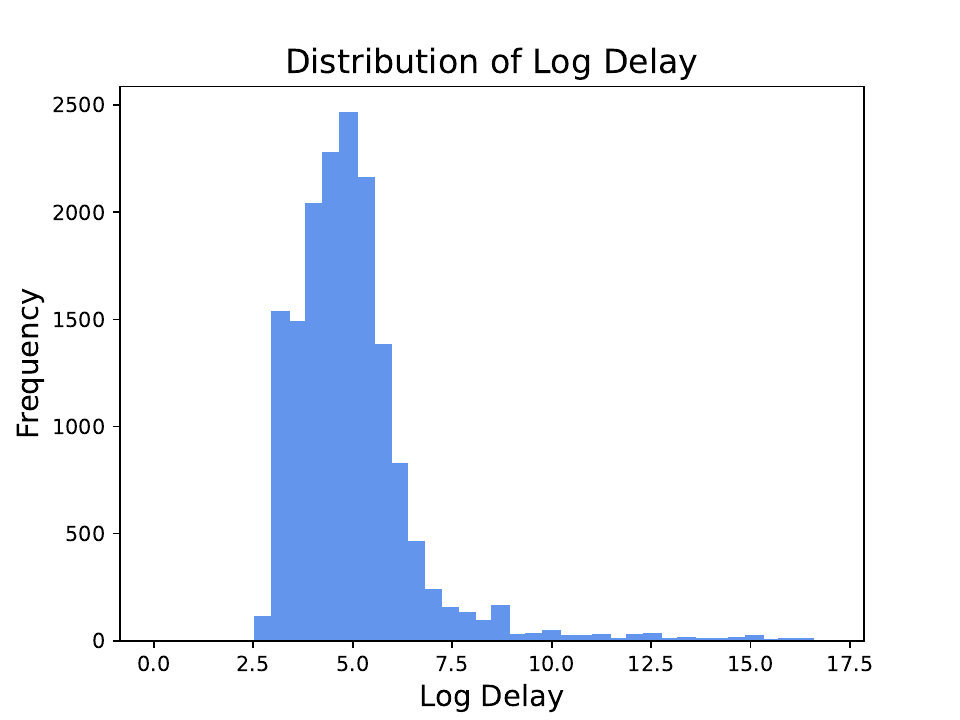}
     \includegraphics[width=0.30\linewidth]{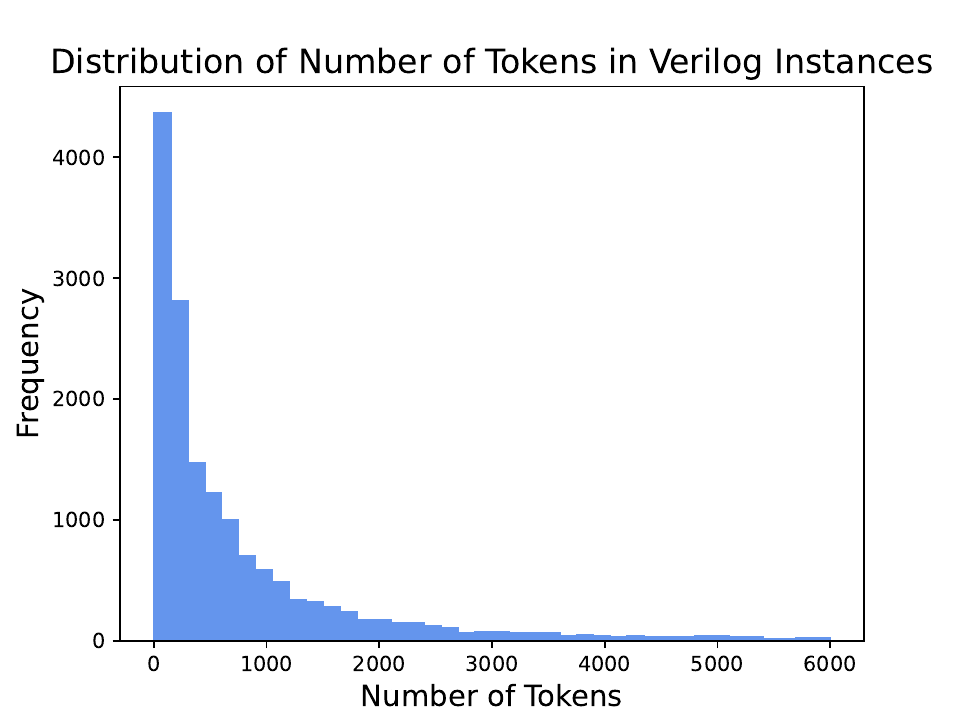}
     \caption{Distribution of labels and the number of tokens in the Verilog dataset.}
     \label{fig:dataset_hist}
\end{figure}

We depict the distribution of labels and the number of tokens in Verilog instances in Figure \ref{fig:dataset_hist}. Verilog data scarcity is a common challenge in developing machine learning tools for RTL level tasks. We note that the majority of Verilog instances contain less than 2000 tokens, with the corresponding circuits having a small area and delay. We note that the labels are obtained under ABC\_AREA=1. However, the label imbalance depicted in Figure 1 is present regardless of the chosen yosys synthesis setup.

\subsection*{B: Results with Different Large Language Models}
\label{appendixA}

\noindent
We employ different version of CodeV; CodeV-Llama, CodeV-DeepSeek and CodeV-CodeQwen, which utilize \texttt{codeLlama-7b}, \texttt{deepseek-coder-6.7b} and \texttt{CodeQwen1.5-7B-Chat} as the base models. These three models are variants of the CodeV model which are fine-tuned on Verilog data. Details about the training procedure can  be found in the original paper. The result in Table \ref{tab:results-other-LLM} are produced with ABC\_AREA=1.

\begin{table*}[h]
\centering
\begin{tabularx}{0.91\linewidth}{lrrrrrrrrr@{}}
\toprule
\multirow{2}{0.5cm}{Approach} & \multicolumn{4}{c}{Area} & \multicolumn{4}{c}{Delay} \\ 
\cmidrule(lr){2-5}\cmidrule(lr){6-9}
& MAE $\downarrow$ & R2 $\uparrow$ & MAPE $\downarrow$ & RSE $\downarrow$ & MAE $\downarrow$ &  R2 $\uparrow$ & MAPE $\downarrow$ & RSE $\downarrow$ \\ \midrule 

CodeV (CodeQwen) + Decoder & 1.070 & 0.563  & 1.975 & 0.437 & 0.732  & 0.368 & 0.139 & 0.632 & \\
CodeV (DeepSeek) + Decoder & 1.061 & 0.566 & 2.184 & 0.434 & 0.738 & 0.367 & 0.143 & 0.633 &  \\
CodeV (CodeLlama) + Decoder & 0.991 & 0.614 & 1.901 & 0.386 & 0.718  & 0.443 & 0.141 & 0.557 & \\\hline
VeriDistill (CodeQwen) & 0.468 & 0.878 & 0.574 & 0.122 & 0.424 & 0.733 & 0.078 & 0.267 &  \\
VeriDistill (DeepSeek)  & 0.484 & 0.875 & 0.622 & 0.125 & 0.426 & 0.706 & 0.077 & 0.294 & \\
VeriDistill (CodeLlama) & 0.495 & 0.862 & 0.629  & 0.138 & 0.415 & 0.728 & 0.076 & 0.272 &  \\
\bottomrule
\end{tabularx}
\caption{The performance of VeriDistill with different Large Language Models.}
\label{tab:results-other-LLM}
\end{table*}

\newpage 

\subsection*{C: Results under Different Synthesis Setting}
\label{appendixD}
To test the robustness of VeriDistill under a different synthesis setting, we re-run synthesis for speed optimization (ABC\_AREA=1 for OpenROAD hyperparameter setting).  We train and evaluate all the methods under the new setup.

\begin{table*}[h]
\centering
\begin{tabularx}{0.81\linewidth}{lrrrrrrrrr@{}}
\toprule
\multirow{2}{0.5cm}{Method} & \multicolumn{4}{c}{Area} & \multicolumn{4}{c}{Delay} \\ 
\cmidrule(lr){2-5}\cmidrule(lr){6-9}
& MAE $\downarrow$ & R2 $\uparrow$ & MAPE $\downarrow$ & RSE $\downarrow$ & MAE $\downarrow$ &  R2 $\uparrow$ & MAPE $\downarrow$ & RSE $\downarrow$ \\ \midrule 
LUT-GNN (Teacher)\hspace{-7pt} & 0.280 & 0.933 & 0.437 & 0.067 & 0.251 & 0.918 & 0.050 & 0.082 &  \\ \hline
\rule{0pt}{2.3ex}AST-XGBoost & 0.773 &  0.745 & 1.494 & 0.362 & 0.521 & 0.632 & 0.096 &  0.565 &\\
AST-GNN & 0.867 & 0.660 & 1.365 & 0.34 & 0.622  & 0.520 & 0.116 & 0.480 \\
AST-GNN w/ KD & 0.898 & 0.670 & 1.327 & 0.33 & 0.654 & 0.561 & 0.122 &  0.439 &  \\ 
CodeV + Decoder & 0.991 & 0.614 & 1.901 & 0.386 & 0.718  & 0.443 & 0.141 & 0.557 & \\
VeriDistill (CodeV) & \textbf{0.495} & \textbf{0.862} & \textbf{0.629}  & \textbf{0.138} & \textbf{0.415} & \textbf{0.728} & \textbf{0.076} & \textbf{0.272} &  \\
\bottomrule
\end{tabularx}
\caption{The performance of different Verilog models on the test dataset under the speed optimization setting. CodeV( CodeV-Llama) is utilized as the base LLM to obtain the above results. }
\label{tab:main-results}
\end{table*}

\subsection*{D: t-SNE Representation of the Last Hidden Layer}
\label{appendixF}

\begin{figure}[h]
    \centering
    \includegraphics[width=0.23\linewidth]{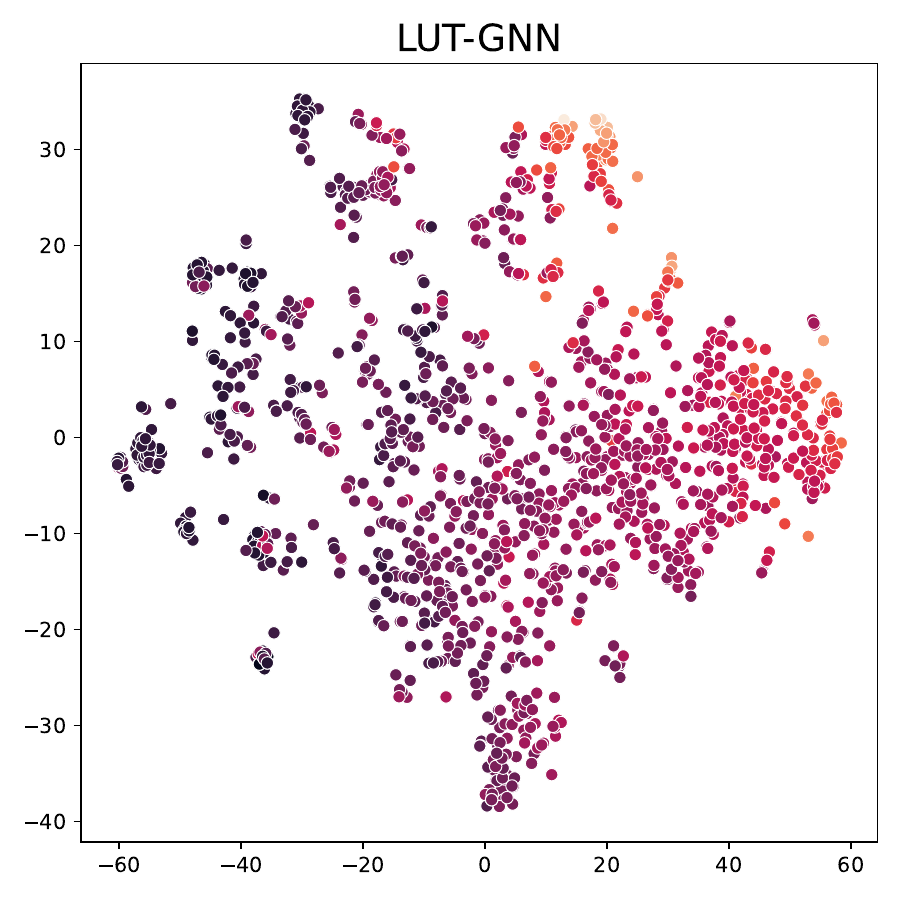}
    \includegraphics[width=0.23\linewidth]{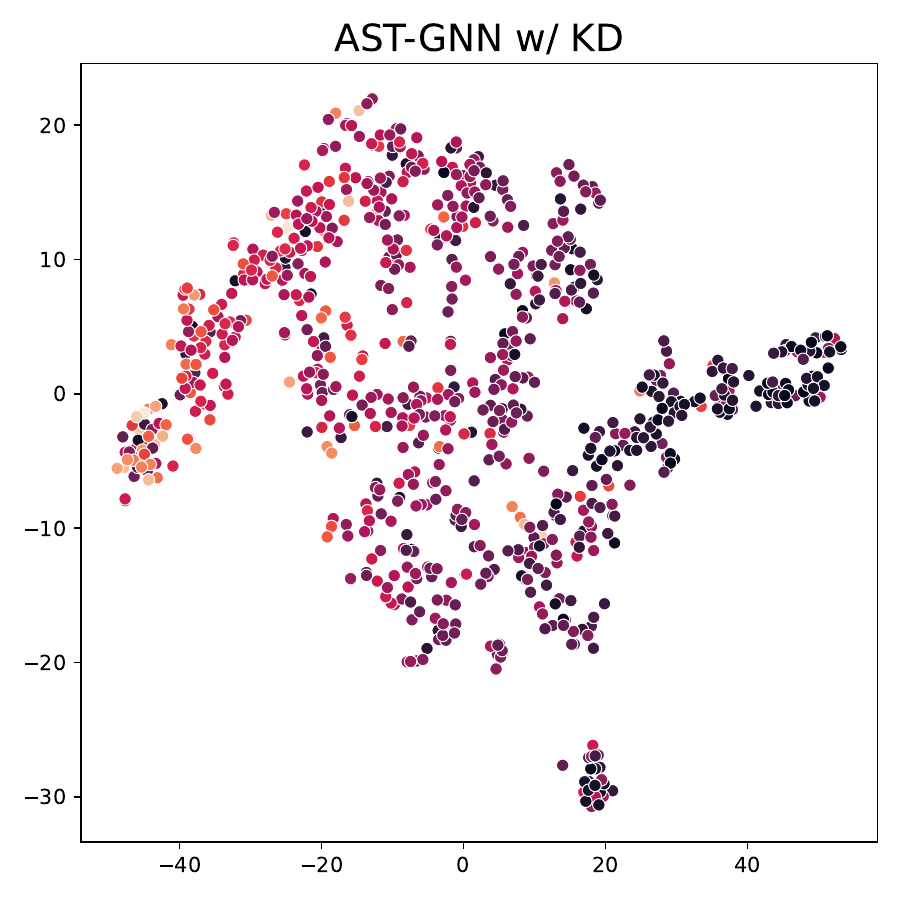}
    \includegraphics[width=0.23\linewidth]{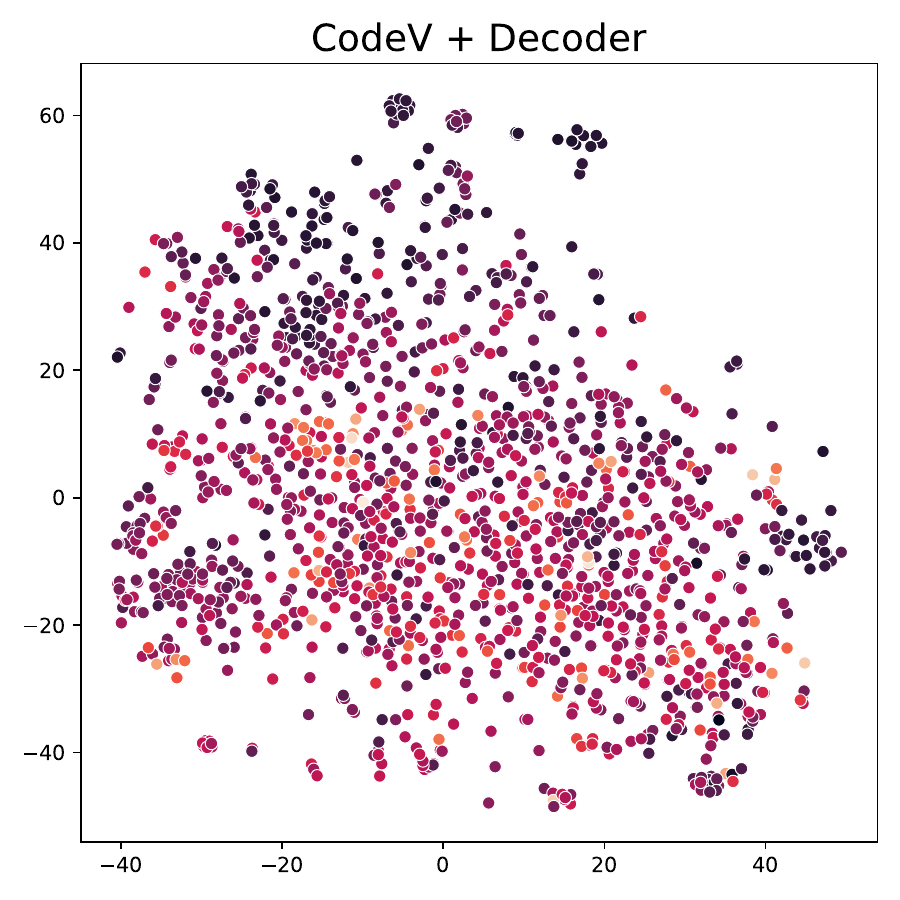}
    \includegraphics[width=0.23\linewidth]{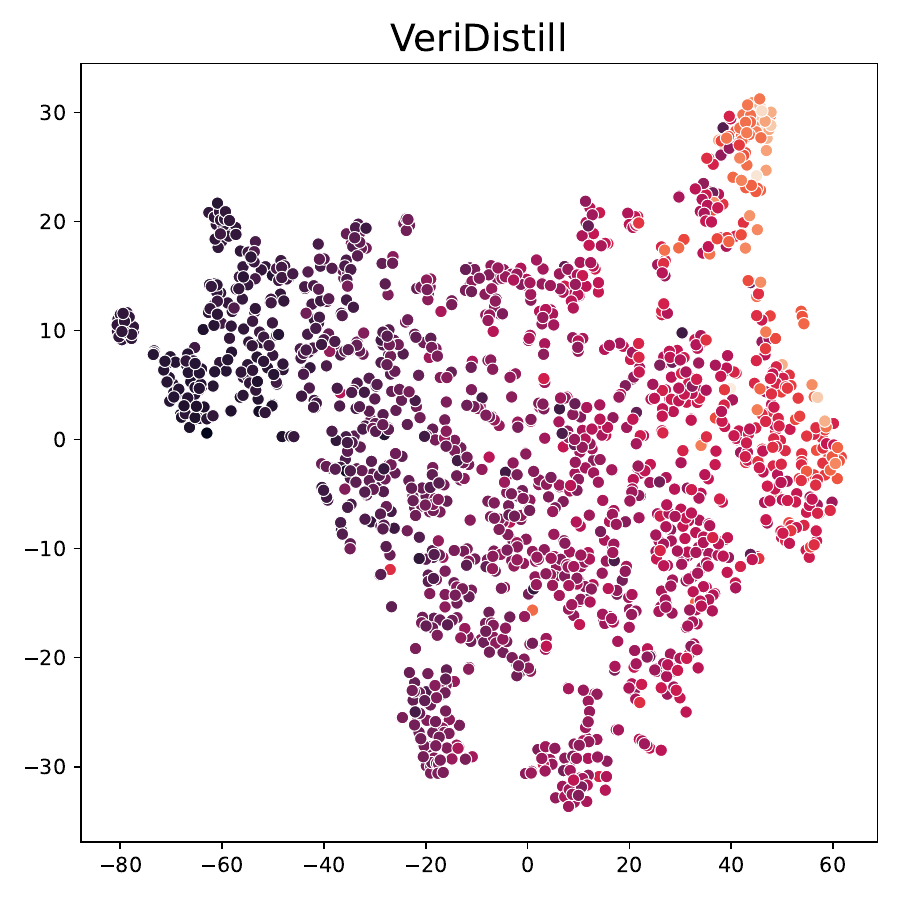}
    \includegraphics[width=0.0352\linewidth]{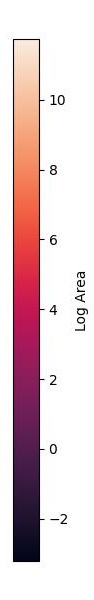}
    \caption{\centering t-SNE representation of the last hidden representation of models on the test data. Color represents the target value (log-area).}
    \label{fig:t-SNE-test}
\end{figure}

In Figure \ref{fig:t-SNE-test}, we present the t-SNE projection of the last hidden space representations on the test data from the teacher model ($Z_{teacher}$) trained for predicting log-area, alongside those from the LLM-based models. As can be seen, the resulting t-SNE representation of the VeriDistill model appears very similar to the one of the LUT-GNN teacher model. Most importantly, the t-SNE of the LUT-GNN model appears to have captured a clear left-to-right pattern in log-area, which shows that the teacher model's  representations have captured a very precise prediction pattern for log-area. This linear pattern has been transferred just as well to VeriDistill. On the contrary, the t-SNE projection of the AST-GNN w/ KD does not exhibit the same vivid pattern as VeriDistill does, where the homogeneity of clusters is abrupter by points of different colors. Finally, the plot of the LLM + Decoder appears much more like an undefined mass, where the log-area values are mixed together indiscriminately.

\subsection*{E: Additional Results on the OpenABCD Benchmark}
\label{appendixC}

The performance of the LUT-GNN (the teacher model) on the OpenABCD benchmark is depicted in Table \ref{tab:teacher-openabcd}. To our supersize, the LLM based methods perform better than the LUT-GNN model on delay prediction. This could be due to the GNN's inability to generalize well to unseen out of distribution circuits.

\begin{table}[H]
\centering
\label{tab:teacher-openabcd}
\begin{tabular}{l|c|c} 
\hline
\multicolumn{1}{l}{}     & \multicolumn{1}{l}{Area}  & Delay   \\ 
\hline
\multicolumn{1}{l}{IP}   & \multicolumn{2}{l}{Absolute Error $\downarrow$}  \\ 
\hline
aes                      & 2.131                     & 2.381   \\
aes\_secworks            & 1.773                     & 0.511   \\
aes\_xcrypt              & 0.946                     & 2.271   \\
des3\_area               & 1.46                      & 1.424   \\
dft                      & 2.859                     & 2.13    \\
dynamic\_node            & 1.786                     & 1.113   \\
ethernet                 & 1.694                     & 1.495   \\
fir                      & 0.099                     & 0.49    \\
fpu                      & 0.126                     & 0.276   \\
i2c                      & 0.472                     & 0.612   \\
idft                     & 2.832                     & 2.098   \\
iir                      & 0.086                     & 1.014   \\
mem\_ctrl                & 0.485                     & 0.509   \\
pci                      & 1.763                     & 1.674   \\
sasc                     & 0.852                     & 0.621   \\
sha256                   & 1.611                     & 1.225   \\
simple\_spi              & 1.151                     & 0.376   \\
spi                      & 0.457                     & 0.714   \\
ss\_pcm                  & 2.051                     & 0.228   \\
tv80                     & 0.117                     & 0.759   \\
usb\_phy                 & 0.108                     & 0.841   \\
vga\_lcd                 & 3.618                     & 1.173   \\
wb\_conmax               & 2.14                      & 0.713   \\ 
\hline
\multicolumn{1}{l}{MEAN} & \multicolumn{1}{l}{1.331} & 1.072   \\
\hline
\end{tabular}
\caption{The performance of LUT-GNN (the teacher model) on the OpenABCD benchmark.}
\end{table}

\noindent
We were able to collect 13 AST graphs from the 23 circuits in the benchmark due to [INSERT REASON]. We outline the performance of different methods on these 13 circuits. As can be seen in Tables 7 and 8, VeriDistill outperforms AST-GNNN based baselines. Interestingly, the LLM (DeepSeek-Coder) based methods perform better than the AST-GNN based baselines, particularly on the delay prediction task.

\begin{table}[H]
\centering
\begin{tabular}{l|c|ccccc}
\multicolumn{1}{c}{}              & \multicolumn{1}{c}{}     & LUT-GNN & LLM + Decoder & VeriDistill    & AST-GNN        & AST-GNN w/ KD  \\ 
\hline
\multicolumn{1}{l}{IP}            & \multicolumn{1}{c}{Area} & \multicolumn{5}{c}{Absolute Error}                                         \\ 
\hline
aes                               & 7.3                      & 2.131   & 2.909         & \textbf{2.746} & 4.045          & 4.284          \\
aes\_secworks                     & 7.629                    & 1.773   & 3.763         & \textbf{0.631} & 23.87          & 1.306          \\
aes\_xcrypt                       & 8.013                    & 0.946   & 3.742         & 3.319          & \textbf{1.664} & 6.101          \\
des3\_area                        & 5.772                    & 1.46    & 1.971         & \textbf{1.068} & 1.674          & 1.818          \\
fir                               & 5.04                     & 0.099   & 2.12          & \textbf{0.829} & 0.884          & 2.545          \\
fpu                               & 7.092                    & 0.126   & 4.077         & \textbf{0.977} & 2.363          & 5.701          \\
i2c                               & 4.193                    & 0.472   & 2.038         & \textbf{1.006} & 1.149          & 2.317          \\
iir                               & 5.508                    & 0.086   & 2.297         & 0.983          & \textbf{0.618} & 3.444          \\
sasc                              & 3.795                    & 0.852   & 0.971         & 0.583          & \textbf{0.054} & 1.879          \\
sha256                            & 6.594                    & 1.611   & 2.259         & \textbf{0.043} & 0.431          & 4.013          \\
spi                               & 5.363                    & 0.457   & 2.321         & 1.968          & \textbf{0.702} & 3.434          \\
ss\_pcm                           & 3.309                    & 2.051   & 0.78          & \textbf{0.192} & 0.432          & 1.424          \\
usb\_phy                          & 3.412                    & 0.108   & 0.359         & \textbf{0.172} & 0.469          & 1.677          \\ 
\hline
\multicolumn{1}{c}{\textbf{MEAN}} & \multicolumn{1}{c}{}     & 0.936   & 2.277         & \textbf{1.117} & 2.95           & 3.073          \\
\hline
\end{tabular}
\caption{The absolute error of predicting the log area on a subset of circuits obtained from the OpenABCD baseline. The highlighted number indicates the best performance across the four baselines (excluding the LUT-GNN model).}
\end{table}

\begin{table}[H]
\centering
\begin{tabular}{l|c|ccccc}
\multicolumn{1}{c}{}              & \multicolumn{1}{c}{}      & LUT-GNN & LLM + Decoder  & VeriDistill    & AST-GNN & AST-GNN w/ KD   \\ 
\hline
\multicolumn{1}{l}{IP}            & \multicolumn{1}{c}{Delay} & \multicolumn{5}{c}{Absolute Error}                                    \\ 
\hline
aes                               & 5.886                     & 2.381   & 0.442          & \textbf{0.212} & 0.369   & 6.684           \\
aes\_secworks                     & 6.33                      & 0.511   & 1.089          & \textbf{0.494} & 7.475   & 21.07           \\
aes\_xcrypt                       & 6.363                     & 2.271   & 0.949          & \textbf{0.713} & 4.042   & 0.411           \\
des3\_area                        & 6.05                      & 1.424   & 0.639          & \textbf{0.22}  & 2.218   & 4.131           \\
fir                               & 5.765                     & 0.49    & 0.83           & 0.527          & 3.298   & \textbf{0.43}   \\
fpu                               & 7.714                     & 0.276   & 2.682          & \textbf{1.574} & 5.635   & 2.205           \\
i2c                               & 5.412                     & 0.612   & 0.692          & \textbf{0.157} & 3.161   & 0.255           \\
iir                               & 5.768                     & 1.014   & 0.736          & 0.572          & 3.141   & \textbf{0.131}  \\
sasc                              & 5.17                      & 0.621   & 0.229          & 0.427          & 3.162   & \textbf{0.102}  \\
sha256                            & 5.587                     & 1.225   & 0.305          & \textbf{0.194} & 2.171   & 0.552           \\
spi                               & 5.82                      & 0.714   & 0.896          & 0.734          & 3.945   & \textbf{0.597}  \\
ss\_pcm                           & 4.89                      & 0.228   & \textbf{0.106} & 0.234          & 3.057   & 1.137           \\
usb\_phy                          & 4.92                      & 0.841   & \textbf{0.05}  & 0.14           & 2.771   & 1.009           \\ 
\hline
\multicolumn{1}{c}{\textbf{MEAN}} & \multicolumn{1}{c}{}      & 0.969 & 0.742          & \textbf{0.477}          & 3.419   & 2.978           \\
\hline
\end{tabular}
\caption{The absolute error of predicting the log delay on a subset of circuits obtained from the OpenABCD baseline. The highlighted number indicates the best performance across the four baselines (excluding the LUT-GNN model).}
\end{table}

\subsection*{F: Training Times}
\label{appendixB}

We outline the time and the number of epochs it takes to train each model under own setup:

\begin{table}[H]
\centering
\begin{tabular}{c|cc}
\hline
Method          & \begin{tabular}[c]{@{}l@{}}Training Time \\ (Till Convergence)\end{tabular} & \begin{tabular}[c]{@{}l@{}}Number of Epochs\\  to Converge\end{tabular} \\ \hline
LUT-GNN         & 21 hours                                                                    & 300                                                                     \\
AST-XGBoost     & 5 minutes                                                                   & N/A                                                                      \\
AST-GNN         & 33 minutes                                                                  & 340                                                                     \\
AST-GNN w/ KD   & 40 minutes                                                                  & 300                                                                     \\
CodeV + Decoder & 12 hours                                                                    & 360                                                                     \\
VeriDistill     & 18 hours                                                                    & 260                                                                     \\ \hline
\end{tabular}
\caption{Training times for the various models.}
\label{tab:resources}
\end{table}

\end{document}